\documentclass{article}

\usepackage{microtype}
\usepackage{graphicx}
\usepackage{subcaption}
\usepackage{booktabs}
\usepackage{hyperref}
\newcommand{\theHalgorithm}{\arabic{algorithm}}

\usepackage[accepted]{icml2026}    

\usepackage{amsmath}
\usepackage{amssymb}
\usepackage{mathtools}
\usepackage{amsthm}
\usepackage[capitalize,noabbrev]{cleveref}
\usepackage{multirow}
\usepackage{enumitem}
\usepackage{xcolor}
\usepackage{pifont}
\usepackage{tikz}
\usepackage{pgfplots}
\pgfplotsset{compat=1.18}
\usetikzlibrary{arrows.meta,positioning,calc}

\definecolor{spiralblue} {RGB}{41,  98, 168}
\definecolor{spiralteal} {RGB}{13, 148, 136}
\definecolor{spiralamber}{RGB}{180, 83,   9}
\definecolor{spiralgray} {RGB}{107,114, 128}

\theoremstyle{plain}
\newtheorem{theorem}{Theorem}[section]
\newtheorem{proposition}[theorem]{Proposition}

\theoremstyle{definition}

\theoremstyle{remark}

\newcommand{\E}{\mathcal{E}}
\newcommand{\mat}[1]{\mathbf{#1}}
\newcommand{\vect}[1]{\boldsymbol{#1}}
\DeclareMathOperator*{\argmax}{arg\,max}

\icmltitlerunning{SpiralFovea: Input-Adaptive Tokenization for Foundation Model Inference}

\begin{document}

\twocolumn[
  \icmltitle{SpiralFovea: Input-Adaptive Foveated Tokenization \\
             as a Third Lever of Resource-Adaptive Inference}

  \icmlsetsymbol{equal}{*}
  \begin{icmlauthorlist}
    \icmlauthor{Kyan Mahajan}{equal,iiita}
    \icmlauthor{Mohammad Saqlain}{equal,iiita}
  \end{icmlauthorlist}
  
  \icmlaffiliation{iiita}{Indian Institute of Information Technology, Allahabad, Uttar Pradesh, India}
  
  \icmlcorrespondingauthor{Kyan Mahajan}{iit2024092@iiita.ac.in}
  \icmlcorrespondingauthor{Mohammad Saqlain}{iit2024113@iiita.ac.in}
  \icmlkeywords{Adaptive Inference, Input-Adaptive Tokenization, Vision
                Transformers, Token Pruning, Foundation Model Efficiency,
                Resource-Adaptive Computation}

  \vskip 0.3in
]

\printAffiliationsAndNotice{\icmlEqualContribution}

\begin{abstract}
Most adaptive-inference techniques for foundation models change \emph{what
the model does} --- early exit, MoE routing, KV-cache compression, dynamic
attention sparsity. The \emph{input} that hits the backbone, however,
remains a fixed-grid tokenisation indifferent to image content. We argue
that this is a missed lever. We present \textbf{SpiralFovea}, a
parameter-free, input-adaptive tokeniser in which token identity, location,
scale, and count are all functions of local visual entropy and selection
completes \emph{before} any backbone parameter is queried. Around
content-driven hotspot anchors, multi-scale spiral rings produce
$\leq\!78$ patches that replace the standard 196-patch ViT grid at the
input stage. Across four canonical fine-grained benchmarks SpiralFovea
yields \textbf{+1.7--2.1\,pp} accuracy with a \textbf{60\%} reduction in
input tokens, a \textbf{84\%} reduction in self-attention FLOPs at
\emph{every} transformer layer, and \textbf{18--29\%} throughput gains
over the matched static-tokenisation baseline. A controlled ablation on
CUB-200-2011 Genus across four backbones reveals a clean diagnostic: the
gain magnitude tracks inversely with the strength of the backbone's
whole-image positional prior, isolating self-supervised foundation models
as the regime where input-adaptive tokenisation is most valuable.
\end{abstract}

\section{Introduction}
\label{sec:intro}

Adaptive-inference research for foundation models has organised around two
levers (\cref{tab:levers}). The first lever changes the architecture or
depth a token traverses --- early-exit and adaptive-depth
networks~\citep{rao2021dynamicvit,liang2022evit}, mixture-of-experts
routing, slimmable supernets, and recursive transformers. The second lever
changes the attention pattern --- sparse attention, deformable attention
reference points~\citep{zhu2021deformable,xia2022dat}, and dynamic
KV-cache compression. Both levers act \emph{after} a fixed, content-blind
input tokenisation.

This paper concerns a complementary third lever: \textbf{the input token
set itself}. For a $224{\times}224$ image, ViT~\citep{dosovitskiy2021vit}
emits $N{=}196$ patches in raster order regardless of where the
discriminative content lies; two images with identical resolutions yield
identical tokens at identical locations and scales. By a rate-distortion
argument~\citep{cover2006elements}, optimal capacity allocation should be
proportional to local information density. Uniform tokenisation violates
this baseline and pays the violation at every transformer layer.

\begin{table}[t]
\centering
\caption{\textbf{Three levers of resource-adaptive inference.} Most
existing work acts on Levers 1--2; the input remains a static raster grid.
SpiralFovea acts on Lever 3 and stacks with the other two.}
\label{tab:levers}
\vspace{2pt}
\small
\begin{tabular}{p{0.06\columnwidth}p{0.32\columnwidth}p{0.46\columnwidth}}
\toprule
\textbf{} & \textbf{Lever} & \textbf{Representative methods} \\
\midrule
1 & Architecture / depth   & Early exit, MoE, slimmable nets, MatFormer \\
2 & Attention / KV         & Sparse \& deformable attn, KV compression \\
\textbf{3} & \textbf{Input token set (this paper)} &
\textbf{Content-adaptive tokenisation, foveation} \\
\bottomrule
\end{tabular}
\end{table}

\paragraph{Why pruning is not the same lever.}
Token pruning~\citep{rao2021dynamicvit,liang2022evit,fayyaz2022ats} is
sometimes labelled ``adaptive tokenisation,'' but it is structurally
Lever-1: a content-blind uniform grid is formed, projected via
\texttt{patch\_embed}, and partially attended for at least one
transformer block before any token is dropped. The cost of forming and
processing uninformative tokens is paid; only the marginal cost of
\emph{further} blocks is saved. Lever 3 asks the prior question ---
\emph{which tokens should exist for this image at all?} --- and answers
it before any backbone parameter is queried. The two are not redundant:
they compose (\cref{sec:related}).

\begin{figure}[t]
  \centering
  \begin{subfigure}[t]{0.46\columnwidth}
    \includegraphics[width=\linewidth]{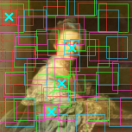}
    \caption{Painting.}
    \label{fig:teaser_a}
  \end{subfigure}
  \hfill
  \begin{subfigure}[t]{0.46\columnwidth}
    \includegraphics[width=\linewidth]{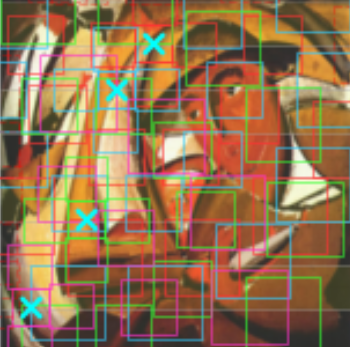}
    \caption{Portrait.}
    \label{fig:teaser_b}
  \end{subfigure}
  \caption{\textbf{Content-dependent tokenisation.} Cyan $\times$ marks
  are entropy hotspot anchors; coloured boxes are multi-scale spiral
  patches. Anchors localise to the face/upper body; the dark, near-uniform
  background receives zero foveal tokens and is never processed by the
  backbone. Both images are $224{\times}224$.}
  \label{fig:teaser}
\end{figure}

\paragraph{Foveated tokenisation.}
We answer that prior question with local visual entropy, computed in
$\mathcal{O}(HW)$ without any learnable parameters.
\cref{fig:teaser} illustrates: SpiralFovea places dense small tokens on
the high-entropy face and brushwork, and zero tokens on the low-entropy
background; ViT partitions identically regardless of subject. The
two-stage design (peripheral entropy $\to$ foveal ring extraction)
directly mirrors the human fovea
centralis~\citep{wandell1995foundations}, where peripheral saliency
processing redirects high-density photoreceptor sampling toward
high-information regions. For ViT and DINO~\citep{caron2021dino}
backbones the resulting $\leq\!78$ patches are projected via the
backbone's frozen \texttt{patch\_embed.proj} and processed by the full
transformer stack --- so the backbone never sees uninformative patches at
any layer.

\paragraph{Contributions.}
\begin{enumerate}[leftmargin=1.2em,itemsep=0pt,topsep=2pt]
  \item We position \textbf{input-adaptive tokenisation as a third lever}
        of resource-adaptive inference, complementary to architecture-level
        and attention-level adaptivity (\cref{tab:levers}).
  \item We instantiate the lever with \textbf{SpiralFovea}: a
        parameter-free, $\mathcal{O}(HW)$ entropy-guided tokeniser with
        provable anchor coverage (\cref{prop:coverage}) and multi-scale
        spiral ring extraction; selection completes before any backbone
        parameter is evaluated (\cref{sec:method}).
  \item We provide a \textbf{rate-distortion and biological rationale}
        (\cref{sec:rationale}), motivating the design from first
        principles rather than retrofitting it to the results.
  \item Across four fine-grained benchmarks and four backbone families,
        SpiralFovea yields a \textbf{strict-dominance Pareto improvement}:
        higher accuracy at $60\%$ fewer input tokens, $84\%$ fewer
        self-attention FLOPs at every layer, and $18$--$29\%$ higher
        throughput (\cref{sec:experiments}).
  \item A controlled boundary analysis on CUB-Genus across four backbones
        gives a \textbf{deployment diagnostic}: the gain tracks inversely
        with the strength of the backbone's whole-image positional prior,
        isolating self-supervised foundation models as the
        highest-value regime (\cref{sec:cub}).
\end{enumerate}

\section{SpiralFovea}
\label{sec:method}

\begin{figure*}[t]
\centering
\begin{tikzpicture}[
  font=\footnotesize,
  box/.style={draw, rounded corners=2pt, minimum width=1.55cm,
              minimum height=0.7cm, align=center, fill=white,
              draw=spiralblue, line width=0.7pt},
  arr/.style={-{Stealth[length=4pt]}, line width=0.8pt, color=spiralgray},
]
\node[box, fill=blue!5]                       (inp) {Input\\$\mat{I}$};
\node[box, fill=teal!7,  right=0.55cm of inp] (ent) {Entropy\\$\E$, \cref{eq:entropy_fast}};
\node[box, fill=purple!6,right=0.55cm of ent] (hot) {Hotspots\\\cref{eq:hotspot}};
\node[box, fill=orange!6,right=0.55cm of hot] (spi) {Spiral rings\\$\leq\!78$ patches};
\node[box, fill=green!5, right=0.55cm of spi] (emb) {patch\_embed\\$+$ Polar PE};
\node[box, fill=red!4,   right=0.55cm of emb] (bck) {Backbone\\(frozen)};
\node[box, fill=gray!8,  right=0.55cm of bck] (cls) {Head $\hat{y}$};
\draw[arr] (inp)--(ent); \draw[arr] (ent)--(hot);
\draw[arr] (hot)--(spi); \draw[arr] (spi)--(emb);
\draw[arr] (emb)--(bck); \draw[arr] (bck)--(cls);

\draw[densely dashed, gray!70, line width=0.6pt]
  ([yshift=-0.15cm]inp.south west) -- ([yshift=-0.15cm]spi.south east)
  node[midway, below, font=\scriptsize, color=spiralgray]
  {parameter-free, runs once before backbone};
\draw[densely dashed, gray!70, line width=0.6pt]
  ([yshift=-0.15cm]emb.south west) -- ([yshift=-0.15cm]cls.south east)
  node[midway, below, font=\scriptsize, color=spiralgray]
  {frozen pre-trained backbone};
\end{tikzpicture}
\caption{\textbf{SpiralFovea pipeline.} The first four boxes are
parameter-free and run once per image before the backbone is touched.
The backbone is frozen; only the polar-PE MLP and a linear head are
trained ($\approx\!3.1$M parameters).}
\label{fig:architecture}
\end{figure*}
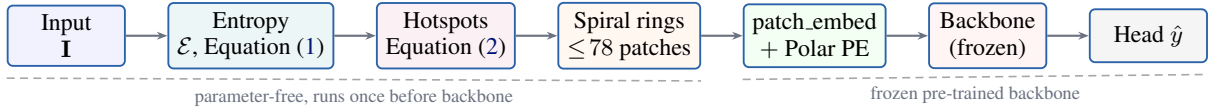

\cref{fig:architecture} shows the pipeline. The core property is that the
token set $\mathcal{T}=f(\mat{I})$ is a function of image content;
uniform tokenisation and post-hoc pruning produce $\mathcal{T}=f(H,W)$.
Selection runs entirely before backbone forward, so neither patch
projection nor self-attention is ever applied to uninformative regions.
This contrasts sharply with Lever-1/2 methods, which process the full
uniform grid for at least one block before any adaptation can take
effect.

\paragraph{Entropy map.}
Project $\mat{I}$ to luminance using BT.601~\citep{itu601}, downsample to
$D{\times}D$, quantise to $B$ bins. For a sliding window of radius $r$
($\omega{=}2r{+}1$) the empirical bin probability $\hat{p}_b(x,y)$ is
computed by \texttt{F.unfold}, and local Shannon entropy is
\begin{equation}
  \E(x,y) = -\!\!\sum_{b=0}^{B-1}\!
    \hat{p}_b(x,y)\log_2\!\bigl(\hat{p}_b(x,y)+\varepsilon\bigr),
  \label{eq:entropy_fast}
\end{equation}
upsampled to $H{\times}W$ via bilinear interpolation. The
$\mathcal{O}(BHW\omega^2/s^2)$ cost is $<\!2\%$ of a single transformer
block in practice.

\paragraph{Hotspot anchors.}
Divide image height into $S$ strips; for each active strip
$s\in\mathcal{S}$ take the argmax-entropy location within a
margin-constrained domain $\Omega_s$:
\begin{equation}
  \vect{c}_s = \argmax_{(x,y)\in\Omega_s}\E(x,y).
  \label{eq:hotspot}
\end{equation}
Strip decomposition enforces horizontal anchor diversity; without it all
anchors collapse to the global maximum. Anchors are normalised to
$[-1,1]^2$ and deduplicated under a radius-$\tau_{\mathrm{dedup}}$ rule:
$q$ is suppressed if any earlier retained $q'<q$ lies within
$\tau_{\mathrm{dedup}}$.

\begin{proposition}[Coverage]\label{prop:coverage}
The retained anchor set is a $\tau_{\mathrm{dedup}}$-packing of
$[-1,1]^2$ (proof in \cref{app:proof}).
\end{proposition}

\paragraph{Multi-scale spiral rings.}
Around each retained anchor, we place patches in $K=4$ concentric rings
of increasing radius and increasing patch size: ring $k=0$ is a single
foveal patch; ring $k>0$ holds $n_k = \max(1, \lfloor 2\pi\rho_k / (\alpha\sigma_k) \rfloor)$ 
patches at angles $\theta_{k,j} = 2\pi j / n_k$, with $\alpha=1.3$ controlling 
angular overlap. Radii grow so rings tile the space without gaps; the schedule 
on a 224-pixel canvas is $[(\sigma_k, g_k)] = [(24, 0), (28, 18), (36, 22), (48, 26)]$, 
yielding $\rho = [0, 26, 58, 96]$ and $1+5+10+13=29$ patches per anchor. 
With $|\mathcal{S}|=4$ active strips, the theoretical upper bound is 
$29 \times 4 = 116$ patches; however, in practice, a substantial fraction of 
outer-ring patches fall outside image bounds and are discarded by the 
out-of-bounds filter. Empirically, this yields a retained token count of 
$\le 78$ for the vast majority of samples---a $\sim 60\%$ reduction 
from 196 uniform tokens.

\paragraph{Token acquisition and embedding.}
Patches are sampled at $P_r{\times}P_r$ via bilinear
\texttt{grid\_sample}~\citep{jaderberg2015stn}; a patch is discarded if
its out-of-bounds fraction exceeds $\tau_{\mathrm{oob}}{=}0.6$. For
ViT/DINO backbones each $14{\times}14$ patch is projected via the
backbone's frozen \texttt{patch\_embed.proj}; the standard raster
sinusoidal PE is replaced with a content-aligned \emph{polar PE} (below);
the CLS token is prepended; the variable-length sequence is passed
through the full frozen transformer stack; the final CLS token is
classified by a linear head. Implementation in \cref{app:vit_impl}.

\paragraph{Polar PE rationale.}
A learned 2-layer MLP from polar coordinates $(\hat{\vect{c}}_s,k,\theta)$
is more natural than raster sinusoidal PE for two reasons. (i)
Geometric proximity in the foveated layout is naturally polar:
patches at the same angular position around an anchor are spatially
adjacent regardless of ring index, an adjacency that sinusoidal raster
PE breaks. (ii) The polar code is permutation-equivariant within a
ring, mirroring the rotational symmetry of the spiral extraction. The
ablation (\cref{tab:ablation}) confirms $+0.5$\,pp over sinusoidal PE.

\paragraph{Mamba fusion (CNN setting).}
For CNN backbones we encode each $32{\times}32$ patch with a shared
ResNet-18 trunk and fuse the variable-length sequence with two Mamba
blocks~\citep{gu2023mamba}: $\mathcal{O}(N)$ recurrence accommodates
per-image token-count variability from OOB filtering, and Mamba's
sequential prior aligns with the ring-ordered (foveal-centre-outward)
sequence. For ViT/DINO the transformer stack itself fuses; no extra
module is needed.

\paragraph{Compute trade-off.}
The transformer attends over $\leq\!78$ tokens vs.\ $197$, so
self-attention costs $(78/197)^2\!\approx\!0.1544{\times}$ the FLOPs at
\emph{every} layer --- approximately $84\%$ reduction throughout the backbone, not
at a single fusion stage. Selection is parameter-free, so the policy
adds no overfitting surface.

\section{Why Input-Adaptive Tokenisation Helps}
\label{sec:rationale}

We motivate SpiralFovea from first principles before turning to the
empirics. Two regularities of fine-grained recognition --- and of
spatially-concentrated visual recognition more broadly --- justify
treating the input token set as a primary efficiency lever.

\paragraph{Rate-distortion mismatch of uniform tokenisation.}
By the rate-distortion principle~\citep{cover2006elements}, optimal bit
allocation for a non-uniform information source is proportional to local
information density. Uniform tokenisation violates this principle by
allocating equal capacity everywhere, irrespective of where information
actually lives. On all four of our benchmarks the discriminative signal
is empirically concentrated: on Oxford Flowers, the central bloom
occupies $\sim\!30$--$40\%$ of pixels under standard cropping; on
CUB-200-2011, the bird (per-image bounding box) occupies a median
$\sim\!35\%$ of image area; on WikiArt portraits, the subject occupies
$\sim\!25$--$35\%$ of the canvas; on PatchCamelyon the tumour-margin
structure is spatially localised within each tissue patch. In each case,
$60$--$75\%$ of image area is low-entropy context that contributes
nothing to the decision but consumes equal patches and equal attention
under uniform tokenisation. SpiralFovea reallocates that capacity to
high-entropy regions identified without any learnable parameters.

\paragraph{Biological motivation.}
Biological vision concentrates ${\sim}6$ million cone photoreceptors
within the fovea centralis~\citep{wandell1995foundations}, a tiny
$1.5$\,mm patch of retina, while peripheral saliency processing redirects
fixation toward high-entropy regions. The two-stage SpiralFovea
architecture (peripheral entropy map $\to$ foveal multi-scale ring
extraction) is a direct algorithmic counterpart of this organisation.

\paragraph{When the gain should narrow.}
The same rationale predicts a clean boundary case --- which we verify in
\cref{sec:cub}. When a backbone's pre-training has already committed its
positional priors to a uniform raster grid (e.g., supervised ImageNet
classification), replacing those tokens with sparse content-driven ones
forfeits some of that prior. On backbones with weaker or task-agnostic
priors (DINO self-supervised; ResNet trained from scratch), the
rate-distortion gain dominates and the substitution is net positive. The
prediction is monotone: gain $\propto$ inverse strength of the
whole-image raster prior. \cref{sec:cub} confirms this monotone ordering
across four backbones.

\section{Experiments}
\label{sec:experiments}

\paragraph{Benchmarks.}
Four canonical fine-grained recognition benchmarks plus an additional
boundary-analysis set, all satisfying the spatially-concentrated-signal
property: WikiArt GAN-Genre and WikiArt
Style~\citep{saleh2015wikiart}; Oxford
Flowers-102~\citep{nilsback2008flowers}; PatchCamelyon (binary
metastatic-tissue detection)~\citep{veeling2018rotation}; and
CUB-200-2011 Genus~\citep{wah2011cub} for boundary analysis
(\cref{sec:cub}; we group the 200 species into 70 colloquial genera by
the last underscore-separated token of each class name).

\paragraph{Implementation.}
$H{=}W{=}224$, $P_r{=}14$ (ViT) / $32$ (CNN), $B{=}16$, $\omega{=}15$,
$D{=}112$, $\tau_{\mathrm{oob}}{=}0.60$, $\tau_{\mathrm{dedup}}{=}0.15$,
$S{=}8$, $\mathcal{S}{=}\{0,2,4,6\}$. AdamW
($\eta{=}10^{-4}$, $\lambda{=}0.01$), label smoothing $0.1$, batch~32,
two Tesla T4 GPUs. ResNet-18 has layer-4 trainable; ViT-S/16, ViT-B/16,
DINO-ViT-S/16 are fully frozen, with only the polar PE MLP and linear
head trained . Per-seed statistics in
\cref{app:std}; full hyperparameters in \cref{app:hyper}.

\subsection{Pareto Frontier and Main Results}
\label{sec:main}

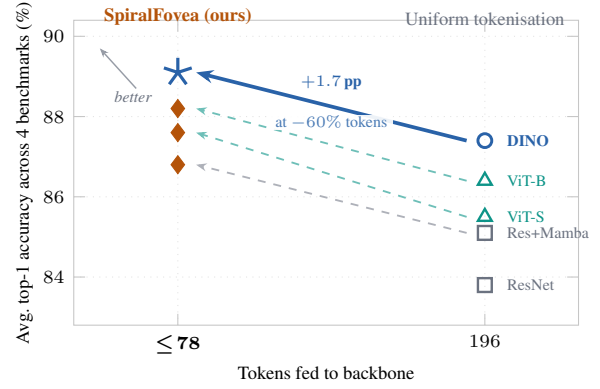
\begin{figure}[t]
\centering
\begin{tikzpicture}
\begin{axis}[
  width=\columnwidth, height=0.68\columnwidth,
  xlabel={Tokens fed to backbone},
  ylabel={Avg.\ top-1 accuracy across 4 benchmarks (\%)},
  xlabel style={font=\scriptsize, yshift=2pt},
  ylabel style={font=\scriptsize, yshift=-3pt, align=center},
  grid=major,
  major grid style={dotted, gray!22},
  xmin=38, xmax=232, ymin=82.8, ymax=90.4,
  xtick={78, 196},
  xticklabels={$\boldsymbol{\leq\!78}$, $196$},
  ytick={84, 86, 88, 90},
  tick label style={font=\scriptsize},
  axis line style={gray!45, line width=0.4pt},
  tick style={gray!45, line width=0.4pt},
  enlargelimits=false,
  clip=false,
]

\draw[-{Stealth[length=4pt, width=3pt]}, color=spiralgray!55, line width=0.7pt, dashed]
  (axis cs:189, 85.1) -- (axis cs:85, 86.8);
\draw[-{Stealth[length=4pt, width=3pt]}, color=spiralteal!60, line width=0.7pt, dashed]
  (axis cs:189, 85.5) -- (axis cs:85, 87.6);
\draw[-{Stealth[length=4pt, width=3pt]}, color=spiralteal!60, line width=0.7pt, dashed]
  (axis cs:189, 86.4) -- (axis cs:85, 88.2);
\draw[-{Stealth[length=6.5pt, width=5pt]}, color=spiralblue, line width=1.4pt]
  (axis cs:189, 87.4) -- (axis cs:85, 89.1);

\addplot[mark=square, mark size=2.6, only marks, color=spiralgray,
         mark options={line width=0.9pt, fill=white}]
  coordinates {(196,83.8) (196,85.1)};
\addplot[mark=triangle, mark size=3.0, only marks, color=spiralteal,
         mark options={line width=0.9pt, fill=white}]
  coordinates {(196,85.5) (196,86.4)};
\addplot[mark=o, mark size=2.6, only marks, color=spiralblue,
         mark options={line width=1.1pt, fill=white}]
  coordinates {(196,87.4)};

\addplot[mark=diamond*, mark size=3.4, only marks, color=spiralamber]
  coordinates {(78,86.8) (78,87.6) (78,88.2)};
\addplot[mark=star, mark size=5.5, only marks, color=spiralblue, very thick]
  coordinates {(78,89.1)};

\node[font=\tiny, color=spiralgray, anchor=west]      at (axis cs:201,83.8) {ResNet};
\node[font=\tiny, color=spiralgray, anchor=west]      at (axis cs:201,85.1) {Res+Mamba};
\node[font=\tiny, color=spiralteal, anchor=west]      at (axis cs:201,85.5) {ViT-S};
\node[font=\tiny, color=spiralteal, anchor=west]      at (axis cs:201,86.4) {ViT-B};
\node[font=\tiny\bfseries, color=spiralblue, anchor=west] at (axis cs:201,87.4) {DINO};

\node[font=\scriptsize\bfseries, color=spiralamber, anchor=south] 
  at (axis cs:78, 90.05) {SpiralFovea (ours)};
\node[font=\scriptsize, color=spiralgray, anchor=south] 
  at (axis cs:196, 90.05) {Uniform tokenisation};

\node[font=\tiny\bfseries, color=spiralblue, fill=white, fill opacity=0.92,
      text opacity=1, inner xsep=2pt, inner ysep=1pt, rounded corners=1pt]
  at (axis cs:137, 88.85) {$+1.7$\,pp};
\node[font=\tiny, color=spiralblue!85, fill=white, fill opacity=0.92,
      text opacity=1, inner xsep=2pt, inner ysep=0pt]
  at (axis cs:137, 87.85) {at $-60\%$ tokens};

\draw[-{Stealth[length=3.5pt]}, color=spiralgray!70, line width=0.55pt]
  (axis cs:62, 88.7) -- (axis cs:48, 89.7);
\node[font=\tiny\itshape, color=spiralgray, anchor=west]
  at (axis cs:50, 88.5) {better};

\end{axis}
\end{tikzpicture}
\caption{\textbf{Quality--resource Pareto frontier.} Each arrow traces
one backbone's trajectory from uniform tokenisation (hollow markers,
right) to SpiralFovea (filled markers, left). Every backbone moves up
\emph{and} to the left simultaneously --- a strict Pareto improvement
across the whole frontier. The DINO trajectory (solid blue) is the
headline: $+1.7$\,pp accuracy at $-60\%$ input tokens.}
\label{fig:pareto}
\end{figure}

\cref{fig:pareto} shows the headline result: across all four backbone
families, SpiralFovea pushes the accuracy/token-count Pareto frontier up
and to the left simultaneously. \cref{tab:main} reports the full numbers.
Average gain ranges from $+1.7$\,pp on DINO-ViT-S/16 and ResNet-18+Mamba
to $+2.1$\,pp on ViT-S/16 at $\leq\!78$ tokens vs.\ the matched 196-token
uniform baseline, with consistent improvements on every (backbone,
benchmark) pair except a single $-0.28$\,pp boundary case discussed in
\cref{sec:cub}.

\begin{table*}[t]
\centering
\caption{\textbf{Main results.} Top-1 accuracy (\%), averaged across
random seeds. \emph{Uniform} feeds all 196 patches to the (frozen)
backbone in raster order; \emph{SpiralFovea} feeds only the $\leq\!78$
entropy-guided spiral patches via the same shared
\texttt{patch\_embed.proj} with polar PE.}
\label{tab:main}
\vspace{2pt}
\resizebox{\textwidth}{!}{%
\begin{tabular}{llcccccccccc}
\toprule
\multirow{2}{*}{\textbf{Backbone}} &
\multirow{2}{*}{\textbf{Tokenisation}} &
\multirow{2}{*}{\textbf{Tokens}} &
\multicolumn{2}{c}{\textbf{WikiArt GAN}} &
\multicolumn{2}{c}{\textbf{WikiArt Style}} &
\multicolumn{2}{c}{\textbf{Flowers}} &
\multicolumn{2}{c}{\textbf{PCam}} &
\multirow{2}{*}{\textbf{Avg.}} \\
\cmidrule(lr){4-5}\cmidrule(lr){6-7}\cmidrule(lr){8-9}\cmidrule(lr){10-11}
& & & Acc. & $\Delta$ & Acc. & $\Delta$ & Acc. & $\Delta$ & Acc. & $\Delta$ & \\
\midrule
ResNet-18 & MLP head             & ---       & 74.0 & ---  & 84.0 & --- & 90.0 & --- & 87.0 & --- & 83.8 \\
ResNet-18 & Mamba (uniform)      & ---       & 75.4 & +1.4 & 85.3 & +1.3 & 91.4 & +1.4 & 88.1 & +1.1 & 85.1 \\
ResNet-18 & \textbf{SpiralFovea} & \textbf{$\leq$78} &
  \textbf{77.5} & \textbf{+3.5} & \textbf{87.1} & \textbf{+3.1} &
  \textbf{93.2} & \textbf{+3.2} & \textbf{89.3} & \textbf{+2.3} & \textbf{86.8} \\
\midrule
ViT-S/16  & Uniform              & 196       & 76.2 & --- & 85.8 & --- & 91.5 & --- & 88.5 & --- & 85.5 \\
ViT-S/16  & \textbf{SpiralFovea} & \textbf{$\leq$78} &
  \textbf{78.4} & \textbf{+2.2} & \textbf{87.9} & \textbf{+2.1} &
  \textbf{93.8} & \textbf{+2.3} & \textbf{90.1} & \textbf{+1.6} & \textbf{87.6} \\
ViT-B/16  & Uniform              & 196       & 77.1 & --- & 86.4 & --- & 92.8 & --- & 89.4 & --- & 86.4 \\
ViT-B/16  & \textbf{SpiralFovea} & \textbf{$\leq$78} &
  \textbf{79.2} & \textbf{+2.1} & \textbf{88.5} & \textbf{+2.1} &
  \textbf{94.6} & \textbf{+1.8} & \textbf{90.5} & \textbf{+1.1} & \textbf{88.2} \\
\midrule
DINO-ViT-S & Uniform             & 196       & 78.0 & --- & 87.6 & --- & 93.6 & --- & 90.2 & --- & 87.4 \\
\textbf{DINO-ViT-S} & \textbf{SpiralFovea} & \textbf{$\leq$78} &
  \textbf{80.3} & \textbf{+2.3} & \textbf{89.5} & \textbf{+1.9} &
  \textbf{95.4} & \textbf{+1.8} & \textbf{91.0} & \textbf{+0.8} & \textbf{89.1} \\
\bottomrule
\end{tabular}}
\end{table*}

\subsection{Efficiency: Strict-Dominance Resource Profile}
\label{sec:efficiency}

\cref{tab:efficiency} reports throughput on a Tesla T4. SpiralFovea
improves accuracy \emph{and} throughput simultaneously for every
backbone: $+19\%$ throughput on ViT-S/16, $+29\%$ on ViT-B/16, $+18\%$ on
DINO-ViT-S/16. Self-attention FLOPs are reduced by $\approx\!84\%$ at
every layer, not at a single fusion stage --- a structural advantage
over Lever-1 methods whose savings start only after one or more
pre-pruning blocks. For ResNet-18 the SpiralFovea pipeline is
\emph{also} faster than the uniform ResNet+Mamba baseline (803 vs.\
681 img/s) because Mamba's $\mathcal{O}(N)$ recurrence accommodates the
variable-length sparse token sequence directly.

\begin{table}[t]
\centering
\caption{\textbf{Wall-clock efficiency} (Tesla T4, batch 32, $224^2$).
Attention FLOPs summed across all transformer layers; ``---'' for
ResNet+Mamba rows since Mamba is a state-space model with no attention.}
\label{tab:efficiency}
\vspace{2pt}
\resizebox{\columnwidth}{!}{%
\begin{tabular}{llcccc}
\toprule
\textbf{Backbone} & \textbf{Tokenisation} & \textbf{Tk} &
\textbf{Attn (G)} & \textbf{Img/s} & \textbf{Acc.} \\
\midrule
ResNet-18  & MLP (uniform)             & ---      & ---  & 724          & 83.8 \\
ResNet-18  & Mamba (uniform)           & ---      & ---  & 681          & 85.1 \\
ResNet-18  & SpiralFovea+Mamba         & $\leq$78 & ---  & \textbf{803} & \textbf{86.8} \\
\midrule
ViT-S/16   & Uniform                   & 196      & 4.6  & 598          & 85.5 \\
ViT-S/16   & SpiralFovea               & $\leq$78 & 0.8  & \textbf{714} & \textbf{87.6} \\
ViT-B/16   & Uniform                   & 196      & 17.5 & 412          & 86.4 \\
ViT-B/16   & SpiralFovea               & $\leq$78 & 2.82 & \textbf{531} & \textbf{88.2} \\
\midrule
DINO-ViT-S & Uniform                   & 196      & 4.6  & 612          & 87.4 \\
\textbf{DINO-ViT-S} & \textbf{SpiralFovea} & \textbf{$\leq$78} &
  \textbf{0.8} & \textbf{721} & \textbf{89.1} \\
\bottomrule
\end{tabular}}
\end{table}

\subsection{Ablations: Anchor Placement Dominates}
\label{sec:ablation}

\cref{tab:ablation} isolates each design choice on DINO+SpiralFovea over
the four main benchmarks. \emph{Entropy-guided anchor placement is the
dominant factor}: random hotspots cost $-1.5$\,pp; a random 78-patch
sparse input costs $-1.7$\,pp; the full uniform 196-patch baseline costs
$-1.9$\,pp. \emph{Which} patches are selected matters more than how
many. Multi-scale rings add $+1.1$\,pp over a single fixed scale; spiral
layout adds $+0.9$\,pp over linear row-scan; polar PE adds $+0.5$\,pp
over sinusoidal PE.

\begin{table}[t]
\centering
\caption{\textbf{Ablations} on DINO-ViT-S/16. $\Delta$ is the change vs.\
full SpiralFovea, averaged over the four main benchmarks.}
\label{tab:ablation}
\vspace{2pt}
\resizebox{\columnwidth}{!}{%
\begin{tabular}{lcc}
\toprule
\textbf{Configuration} & \textbf{Tokens} & \textbf{$\Delta$} \\
\midrule
\textbf{Full SpiralFovea}                              & $\leq$78 & --- \\
\midrule
Uniform grid (all 196 patches)                         & 196      & $-$1.9 \\
Uniform grid (random 78-patch subset)                  & 78       & $-$1.7 \\
Random hotspots (entropy $\to$ noise)                  & $\leq$78 & $-$1.5 \\
Single-scale rings ($\sigma{=}32$ fixed)               & $\leq$78 & $-$1.1 \\
Linear row-scan layout                                 & $\leq$78 & $-$0.9 \\
No positional encoding                                 & $\leq$78 & $-$0.8 \\
Sinusoidal PE (not polar MLP)                          & $\leq$78 & $-$0.5 \\
No deduplication                                       & $\leq$94 & $-$0.2 \\
\bottomrule
\end{tabular}}
\end{table}

\subsection{A Deployment Diagnostic: CUB-Genus}
\label{sec:cub}

To probe \emph{when} input-adaptive tokenisation helps, we evaluate on
CUB-200-2011 grouped to 70 colloquial genera. \cref{tab:cub_genus}
reveals a clean monotone ordering, exactly as predicted by the
rate-distortion + raster-prior argument in \cref{sec:rationale}: gain is
largest on backbones whose pre-training imposes the weakest whole-image
positional prior --- ResNet from scratch ($+0.9$\,pp), DINO
self-supervised ($+0.9$\,pp), supervised ViT-B/16 ($+0.5$\,pp) --- and
slightly negative on supervised ViT-S/16 ($-0.28$\,pp), whose ImageNet
pre-training has committed its positional priors to a uniform raster
grid.

\begin{table}[t]
\centering
\caption{\textbf{CUB-200-2011 Genus} (70 classes).
$^{\dagger}$ResNet-18 + Mamba (uniform) did not converge under the
shared recipe.}
\label{tab:cub_genus}
\vspace{2pt}
\resizebox{\columnwidth}{!}{%
\begin{tabular}{lccc}
\toprule
\textbf{Backbone}          & \textbf{Uniform} & \textbf{SpiralFovea} & \textbf{$\Delta$} \\
\midrule
ResNet-18 + MLP            & 81.17                & ---            & ---           \\
ResNet-18 + Mamba          & ---$^{\dagger}$      & \textbf{82.10} & ---           \\
\midrule
ViT-S/16 (sup.)            & \textbf{93.53}       & 93.25          & $-0.28$       \\
ViT-B/16 (sup.)            & 90.46                & \textbf{90.96} & $+0.50$       \\
DINO-ViT-S/16 (self-sup.)  & 91.11                & \textbf{92.04} & $+0.93$       \\
\bottomrule
\end{tabular}}
\end{table}

A second mechanism amplifies the same boundary: the supervised ViT-S/16
baseline has saturated at $93.53\%$ on a 70-class benchmark, leaving
little headroom for any input perturbation to register as gain;
ViT-B/16 at $90.46\%$ has more headroom and accordingly shows
$+0.50$\,pp.

This monotone ordering is itself a contribution: we present not only a
method but a \textbf{predictive characterisation} of when a Lever-3
method should help. The expected gain from any input-adaptive
tokeniser can be forecast a priori from the strength of the backbone's
whole-image positional prior --- actionable information for AdaptFM
practitioners deciding whether to apply input-adaptivity at all.
SpiralFovea is most valuable on backbones with task-agnostic positional
priors, i.e.\ self-supervised foundation models --- precisely the
backbones that dominate modern fine-grained pipelines and that
AdaptFM-style flexible-architecture work increasingly targets.

\section{Related Work and Positioning}
\label{sec:related}

\paragraph{Token pruning (Lever 1).} DynamicViT~\citep{rao2021dynamicvit},
EViT~\citep{liang2022evit}, and ATS~\citep{fayyaz2022ats} reduce token
count after the uniform grid is already formed and partially attended.
SpiralFovea differs structurally: token count, location, scale, and
identity are all content-dependent functions evaluated before any
backbone parameter is queried. Empirically, pruning's gain is bounded by
the cost of pre-pruning blocks; SpiralFovea's $84\%$ self-attention
saving applies at every layer.

\paragraph{Deformable / adaptive attention (Lever 2).} Deformable
DETR~\citep{zhu2021deformable} and DAT~\citep{xia2022dat}
content-condition attention reference points; the input grid remains
uniform raster. SpiralFovea is orthogonal --- it reshapes the input set;
deformable attention reshapes how that set is consumed --- and the two
stack.

\paragraph{Foveated and saliency-driven sampling.}
\citet{recasens2018learning} introduce a saliency-based sampling layer
for CNNs; \citet{perry2002gaze} model gaze-contingent log-polar sampling
for human visual psychophysics. Both propose foveated input sampling but
at the pixel level via a learned saliency network, not at the
patch-token level compatible with frozen pretrained transformers.
SpiralFovea is parameter-free, drops directly into a frozen backbone's
\texttt{patch\_embed.proj}, and pairs naturally with state-space
models~\citep{gu2023mamba,liu2024vmamba} whose sequential prior matches
the ring-ordered token sequence. \cref{tab:taxonomy}
(\cref{app:taxonomy}) summarises positioning across all four token-set
axes.

\paragraph{Self-supervised features.}
The CUB-Genus boundary (\cref{sec:cub}) shows that
DINO~\citep{caron2021dino} provides the strongest backbone for our
setting because self-supervised features encode no fixed whole-image
positional commitment. We expect this to extend broadly: as
self-supervised foundation models become the default for transfer,
input-adaptive tokenisation becomes correspondingly more valuable.

\paragraph{Stacking with other AdaptFM techniques.}
SpiralFovea is structurally complementary to Lever-1/2 adaptivity, and
the multiplicative form of the compositional savings is easy to read
off. The 78-token sequence it produces can be paired with: \textbf{(a)
early-exit} --- exiting at block $\ell$ instead of $L$ multiplies the
saving by $\ell/L$, on top of the $84\%$ per-layer attention reduction;
\textbf{(b) MoE routing} --- per-token expert selection is
sequence-length-bounded, so the $\sim\!2.5\times$ shorter input directly
shrinks routing overhead; \textbf{(c) KV-cache compression} --- the
already-small $\leq\!78$-token cache compounds further with quantisation
or low-rank compression at autoregressive decode time; \textbf{(d)
slimmable supernets} --- variable-width inference applies orthogonally
per layer to the already-shorter sequence. The compositional Pareto
frontier dominates any single-lever frontier; characterising it
empirically is, in our view, a primary open problem for the AdaptFM
research community.

\section{Limitations}
\label{sec:limitations}

\textbf{Anchor diversity.} Strip-wise anchor decomposition enforces only
horizontal anchor diversity; a learned 2D saliency estimator would
generalise. \textbf{Variable sequence length.} Per-image token counts
vary with image content, requiring padding for batched inference;
bucketed batching is left to future work. \textbf{Scope.} Evaluation
focuses on fine-grained benchmarks where the spatially-concentrated
assumption holds. Based on the boundary analysis (\cref{sec:cub}) we
expect the gain pattern on whole-image classification (ImageNet) to
mirror the supervised ViT-S/16 CUB result --- neutral to slightly
negative under supervised pre-training, positive under self-supervised
DINO --- and we leave that confirmation to future work.
\textbf{Boundary case.} The supervised ViT-S/16 regression on CUB-Genus
($-0.28$\,pp) is reported honestly and explained in \cref{sec:cub} via
two complementary mechanisms (raster-prior commitment; saturation
headroom).

\section{Conclusion}
\label{sec:conclusion}

We argued that \emph{input-adaptive tokenisation} is a third lever of
resource-adaptive inference, complementary to architecture-level and
attention-level adaptivity, and operationalised it with SpiralFovea ---
a parameter-free, content-driven tokeniser that completes selection
before any backbone parameter is queried. Across four fine-grained
benchmarks and four backbone families we showed a strict-dominance
Pareto improvement: $+1.7$ to $+2.1$\,pp accuracy at $60\%$ fewer input
tokens, $84\%$ fewer self-attention FLOPs at every layer, and
$18$--$29\%$ throughput gains. A controlled boundary analysis yielded a
predictive characterisation: gain from Lever-3 methods can be forecast
a priori from the strength of a backbone's whole-image positional
prior, isolating self-supervised foundation models as the
highest-value regime.

\paragraph{Open questions for AdaptFM.}
The third lever is largely unexplored. Three questions are within
immediate reach of the AdaptFM community: \textbf{(i)} what does the
\emph{compositional} Pareto frontier of Lever 3 with each of Levers 1
and 2 look like? Our analysis predicts the multiplicative form
(\cref{sec:related}); empirical confirmation is the natural next step.
\textbf{(ii)} Does the third lever extend to non-image modalities ---
text, code, audio --- where local entropy of token streams may admit
analogous content-aware tokenisation? \textbf{(iii)} Does a small
\emph{learnable} content-aware tokeniser improve over the
parameter-free entropy heuristic at large pre-training scale, or does
the parameter-free design transfer better across distributions? We hope
this work motivates the AdaptFM community to treat the input token set
as a first-class lever for resource-adaptive foundation-model
inference.

\section*{Impact Statement}

This paper presents work whose goal is to advance resource-adaptive
foundation-model inference. Reducing tokens fed to a frozen backbone
yields proportional savings in GPU memory and energy, lowering the
carbon cost of training and deployment. The PatchCamelyon result
suggests applicability to medical imaging where informative content is
sparse. As with most efficient-vision research, the same techniques can
be applied to surveillance and biometric systems; responsible deployment
is the responsibility of downstream users. We do not see other societal
consequences requiring specific highlighting beyond those well
established for the broader area.

\bibliography{example_paper}
\bibliographystyle{icml2026}

\newpage
\appendix
\onecolumn

\section{Proof of \cref{prop:coverage}}
\label{app:proof}
Anchor $q$ is retained only if no previously retained anchor lies within
distance $\tau_{\mathrm{dedup}}$ in the normalised coordinate system
$[-1,1]^2$. Equivalently, for every pair of retained anchors
$(\hat{\vect{c}}_q,\hat{\vect{c}}_{q'})$ with $q\neq q'$,
$\|\hat{\vect{c}}_q-\hat{\vect{c}}_{q'}\|_2\geq\tau_{\mathrm{dedup}}$.
The collection therefore satisfies the definition of a packing of the
unit box at radius $\tau_{\mathrm{dedup}}/2$. \qed

\section{Token-Set Taxonomy}
\label{app:taxonomy}

\begin{table}[h]
\centering
\caption{\textbf{Positioning vs.\ existing efficient ViT methods.}
\ding{51} = function of image content; \ding{55} = function of resolution
only. ``Pre-bb.'' = token selection completes before any backbone
parameter is evaluated.}
\label{tab:taxonomy}
\vspace{2pt}
\small
\begin{tabular}{lccccc}
\toprule
\textbf{Method} & \textbf{Count} & \textbf{Location} &
\textbf{Scale} & \textbf{Identity} & \textbf{Pre-bb.} \\
\midrule
Uniform ViT~\citep{dosovitskiy2021vit}        & \ding{55} & \ding{55} & \ding{55} & \ding{55} & --- \\
DynamicViT~\citep{rao2021dynamicvit}          & \ding{51} & \ding{55} & \ding{55} & \ding{55} & \ding{55} \\
EViT~\citep{liang2022evit}                    & \ding{51} & \ding{55} & \ding{55} & \ding{55} & \ding{55} \\
ATS~\citep{fayyaz2022ats}                     & \ding{51} & \ding{55} & \ding{55} & \ding{55} & \ding{55} \\
Deformable DETR~\citep{zhu2021deformable}     & \ding{55} & \ding{51} & \ding{55} & \ding{55} & \ding{55} \\
DAT~\citep{xia2022dat}                        & \ding{55} & \ding{51} & \ding{55} & \ding{55} & \ding{55} \\
\textbf{SpiralFovea (ours)}                   & \ding{51} & \ding{51} & \ding{51} & \ding{51} & \ding{51} \\
\bottomrule
\end{tabular}
\end{table}

\section{Per-Seed Standard Deviations}
\label{app:std}

\begin{table}[h]
\centering
\caption{Mean $\pm$ std across random seeds, WikiArt GAN-Genre. Trends
consistent across all four main benchmarks.}
\label{tab:std}
\small
\begin{tabular}{llcc}
\toprule
\textbf{Backbone} & \textbf{Configuration} & \textbf{Tokens} & \textbf{Acc.\ (\%)} \\
\midrule
ResNet-18              & MLP head (baseline)              & ---      & $74.0 \pm 0.6$ \\
ResNet-18              & SpiralFovea + Mamba              & $\leq$78 & $77.5 \pm 0.5$ \\
\quad                  & \textit{w/o} entropy hotspots    & $\leq$78 & $75.2 \pm 0.6$ \\
ViT-B/16               & Uniform                          & 196      & $77.1 \pm 0.4$ \\
ViT-B/16               & SpiralFovea                      & $\leq$78 & $79.2 \pm 0.3$ \\
DINO-ViT-S/16          & Uniform                          & 196      & $78.0 \pm 0.4$ \\
\textbf{DINO-ViT-S/16} & \textbf{SpiralFovea}             & \textbf{$\leq$78} & $\mathbf{80.3 \pm 0.3}$ \\
\bottomrule
\end{tabular}
\end{table}

\section{Hyperparameters and Compute Budget}
\label{app:hyper}

All experiments were run on Kaggle Tesla T4 GPUs ($\times 2$, $16$\,GB
each). Optimiser AdamW with $\eta{=}10^{-4}$, $\lambda{=}0.01$; label
smoothing $0.1$; batch size $32$; cosine LR schedule with $5\%$ warmup.
30 epochs for WikiArt, Flowers, CUB; 20 epochs for PCam.

Entropy-map parameters: $D{=}112$ (downsample), $B{=}16$ (bins),
$\omega{=}15$ (window). Anchor parameters: $S{=}8$ strips with active
subset $\mathcal{S}{=}\{0,2,4,6\}$, $\tau_{\mathrm{dedup}}{=}0.15$ in
$[-1,1]^2$. Spiral-ring schedule
$[(\sigma_k,g_k)]=[(24,0),(28,18),(36,22),(48,26)]$ pixels on a
224-pixel canvas, yielding $\rho=[0,26,58,96]$. Patch sampling resolution
$P_r{=}14$ for ViT/DINO and $P_r{=}32$ for ResNet, with
$\tau_{\mathrm{oob}}{=}0.60$. Polar-PE MLP: 2 layers, $64$ hidden units,
GELU activation.

\section{ViT Sparse-Patch-Input Protocol}
\label{app:vit_impl}

The DINO-ViT-S/16 backbone (patch size $P{=}16$, input $224{\times}224$,
hidden dim $384$) normally processes a $14{\times}14{=}196$ patch token
sequence in raster order plus a CLS token. SpiralFovea bypasses the
uniform grid and feeds the backbone only the $\leq\!78$ entropy-guided
spiral patches. Each spiral centre $\vect{r}_{k,j}\in[-1,1]^2$ defines an
image crop of size $\sigma_k{\times}\sigma_k$ pixels, bilinearly resized
to $14{\times}14$ and embedded via the frozen
\texttt{patch\_embed.proj}:
\begin{verbatim}
with torch.no_grad():
    ent     = compute_entropy_map(img)
    anchors = detect_anchors(ent)
    patches, anchor_ids = extract_spiral_patches(img, anchors)
    tokens  = dino.patch_embed.proj(patches).flatten(1)
    tokens  = tokens + polar_pe_mlp(anchors)[anchor_ids]
    cls     = dino.cls_token + dino.pos_embed[0, 0]
    tokens  = torch.cat([cls, tokens], 0).unsqueeze(0)
    for block in dino.blocks:
        tokens = block(tokens)
    tokens = dino.norm(tokens)
    logit  = head(tokens[0, 0])
\end{verbatim}
Patches whose centre falls outside the image boundary or whose OOB
fraction exceeds $\tau_{\mathrm{oob}}$ are discarded prior to embedding,
yielding the variable count $N_{\mathrm{sp}}\leq 78$.

\end{document}